%% file: main.tex
\documentclass[10pt,twocolumn,letterpaper]{article}

\usepackage[pagenumbers]{iccv} 

\usepackage{colortbl}
\usepackage{arydshln}
\usepackage{multirow}
\usepackage{algorithmic}
\usepackage{algorithm}


\definecolor{iccvblue}{rgb}{0.21,0.49,0.74}
\usepackage[pagebackref,breaklinks,colorlinks,allcolors=iccvblue]{hyperref}

\def\paperID{7967} 
\def\confName{ICCV}
\def\confYear{2025}

\title{MVQA: Mamba with Unified Sampling for Efficient Video Quality Assessment}

\author{Yachun Mi$^{1}$ \qquad Yu Li$^{1}$ \qquad Weicheng Meng$^1$ \qquad Chaofeng Chen$^2$  \qquad \\ Chen Hui$^1$  \qquad  Shaohui Liu\footnotemark[1]~$^1$ \\ $^1$ Harbin Institute of Technology \qquad  $^2$ Sun Yat-sen University
}

\begin{document}
\maketitle
\input{0_abstract}    
\input{1_introduction}

\input{2_related}

\input{3_approach}

\input{4_experiments}
\input{5_conclusion}

{
    \small
    \bibliographystyle{ieeenat_fullname}
    \bibliography{main}
}

\end{document}

%% file: 0_abstract.tex
\begin{abstract}
	
The rapid growth of long-duration, high-definition videos has made efficient video quality assessment (VQA) a critical challenge.
Existing research typically tackles this problem through two main strategies: reducing model parameters and resampling inputs.
However, light-weight Convolution Neural Networks (CNN) and Transformers often struggle to balance efficiency with high performance due to the requirement of long-range modeling capabilities. 
Recently, the state-space model, particularly Mamba, has emerged as a promising alternative, offering linear complexity with respect to sequence length. 
Meanwhile, efficient VQA heavily depends on resampling long sequences to minimize computational costs, yet current resampling methods are often weak in preserving essential semantic information.
In this work, we present MVQA, a Mamba-based model designed for efficient VQA along with a novel Unified Semantic and Distortion Sampling (USDS) approach.
USDS combines semantic patch sampling from low-resolution videos and distortion patch sampling from original-resolution videos. The former captures semantically dense regions, while the latter retains critical distortion details. 
To prevent computation increase from dual inputs, we propose a fusion mechanism using pre-defined masks, enabling a unified sampling strategy that captures both semantic and quality information without additional computational burden.
Experiments show that the proposed MVQA, equipped with USDS, achieve comparable  performance to state-of-the-art methods while being $2\times$ as fast and requiring only $1/5$ GPU memory.

\end{abstract}

%% file: 1_introduction.tex
\section{Introduction}
\label{sec:intro}

\begin{figure}[htbp]
	\centering
	\setlength{\abovecaptionskip}{0.1cm}
	\begin{minipage}{0.49\linewidth}
		\centering
		\includegraphics[width=1\linewidth]{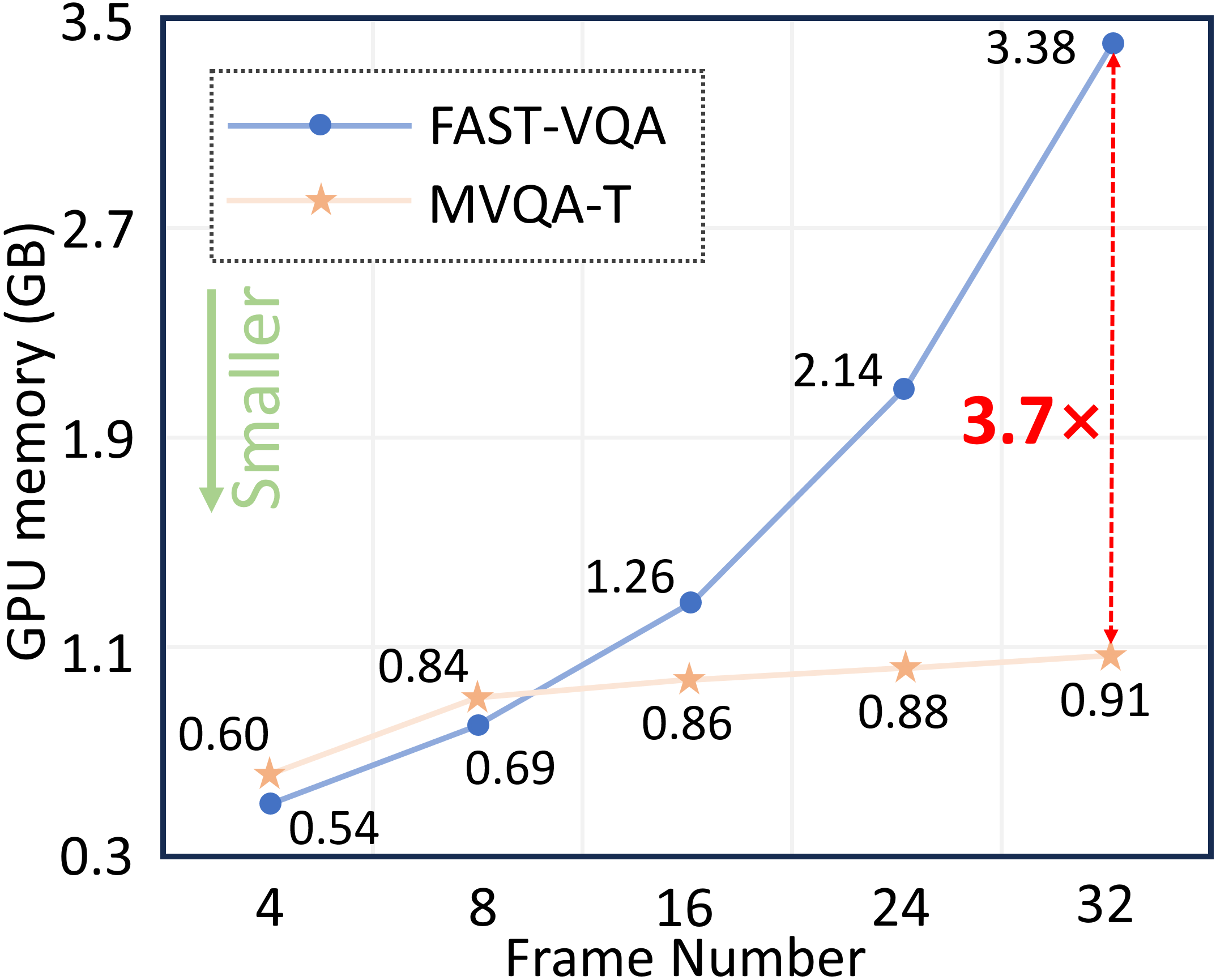}
		\centerline{ \footnotesize (a) GPU Memory Comparison}
	\end{minipage}
	\begin{minipage}{0.49\linewidth}
		\centering
		\includegraphics[width=1\linewidth]{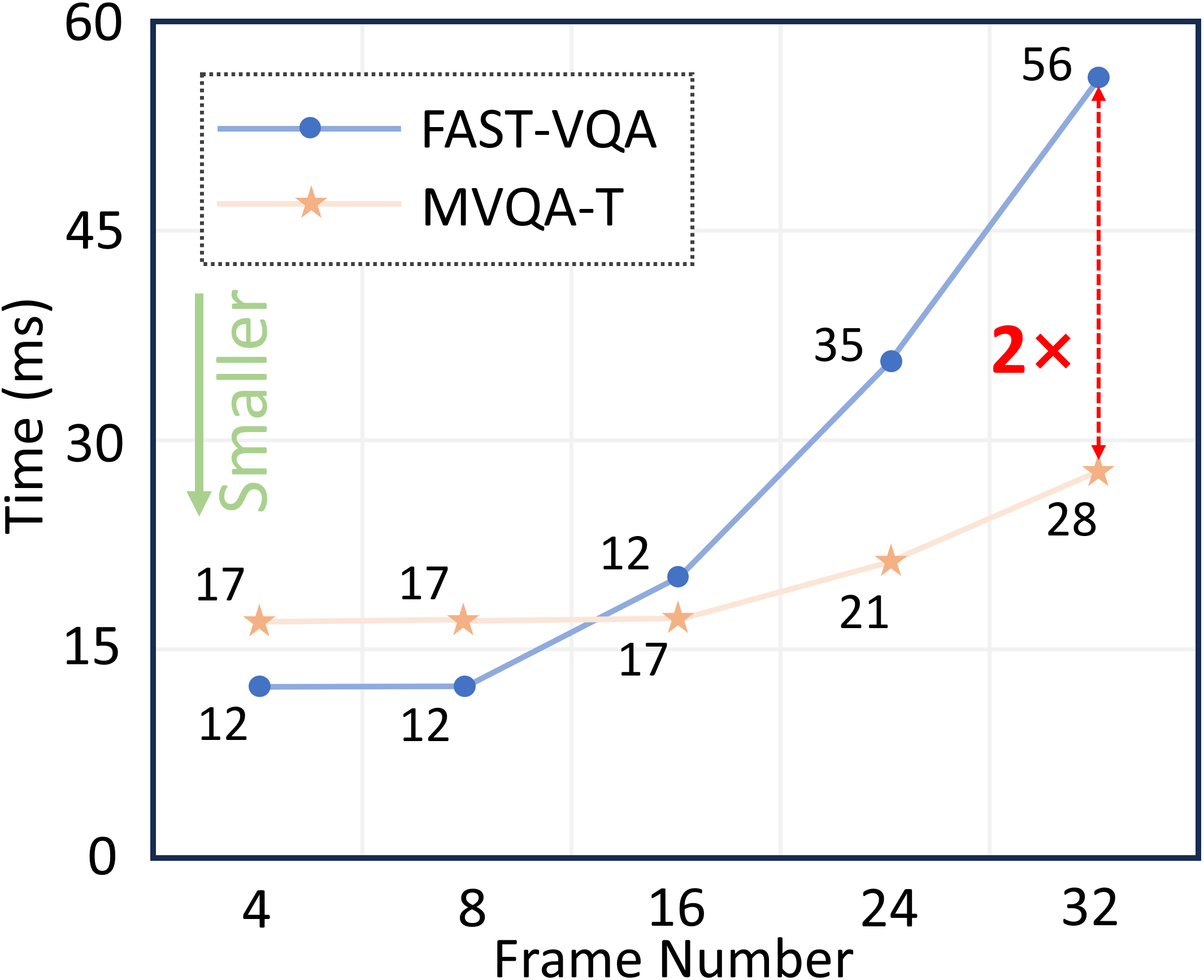}
		\centerline{\footnotesize (b) Speed Comparison}
	\end{minipage}
	\begin{minipage}{0.49\linewidth}
		\centering
		\includegraphics[width=1\linewidth]{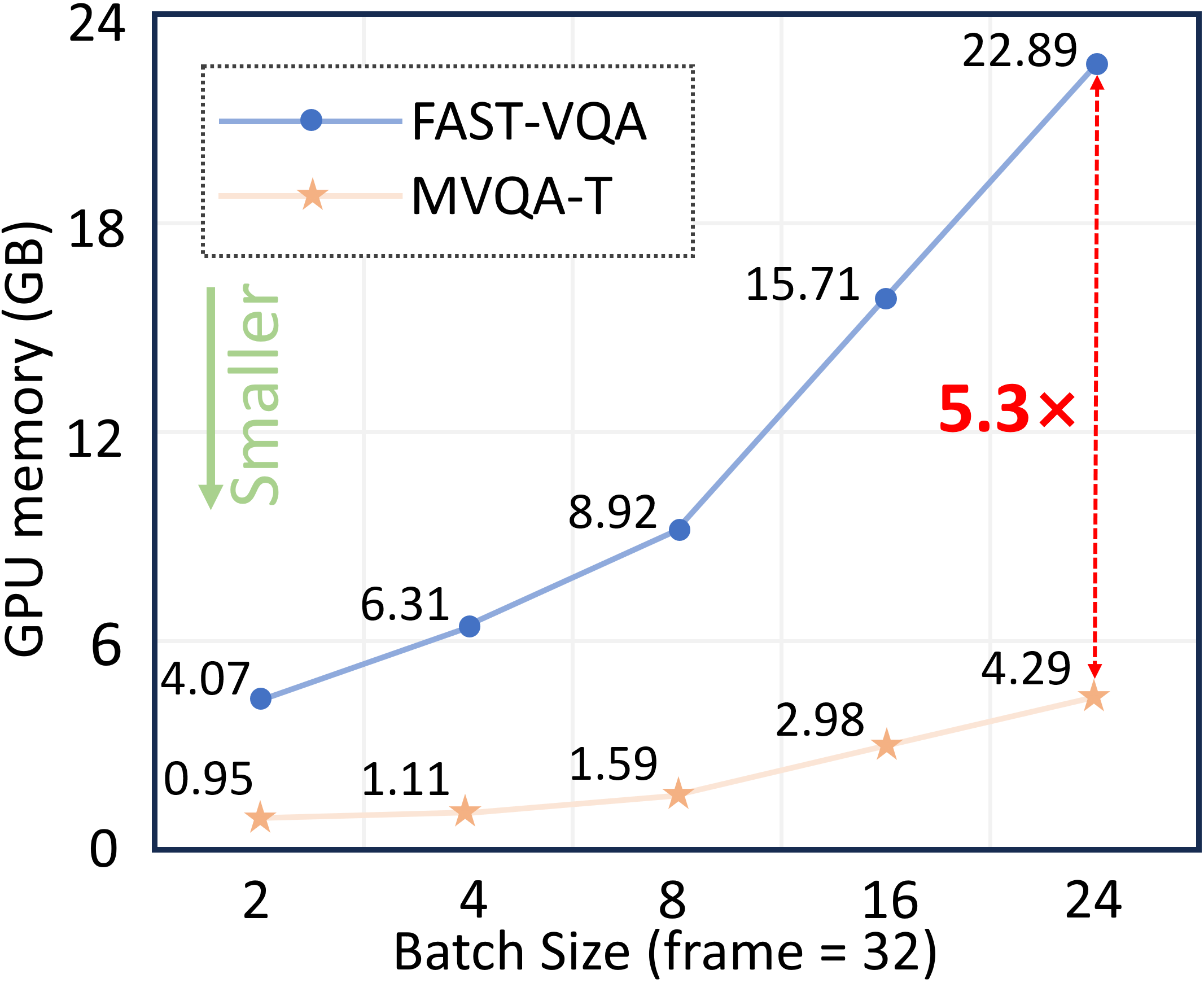}
		\centerline{\footnotesize (c) GPU Memory Comparison}
	\end{minipage}
	\begin{minipage}{0.49\linewidth}
		\centering
		\includegraphics[width=1\linewidth]{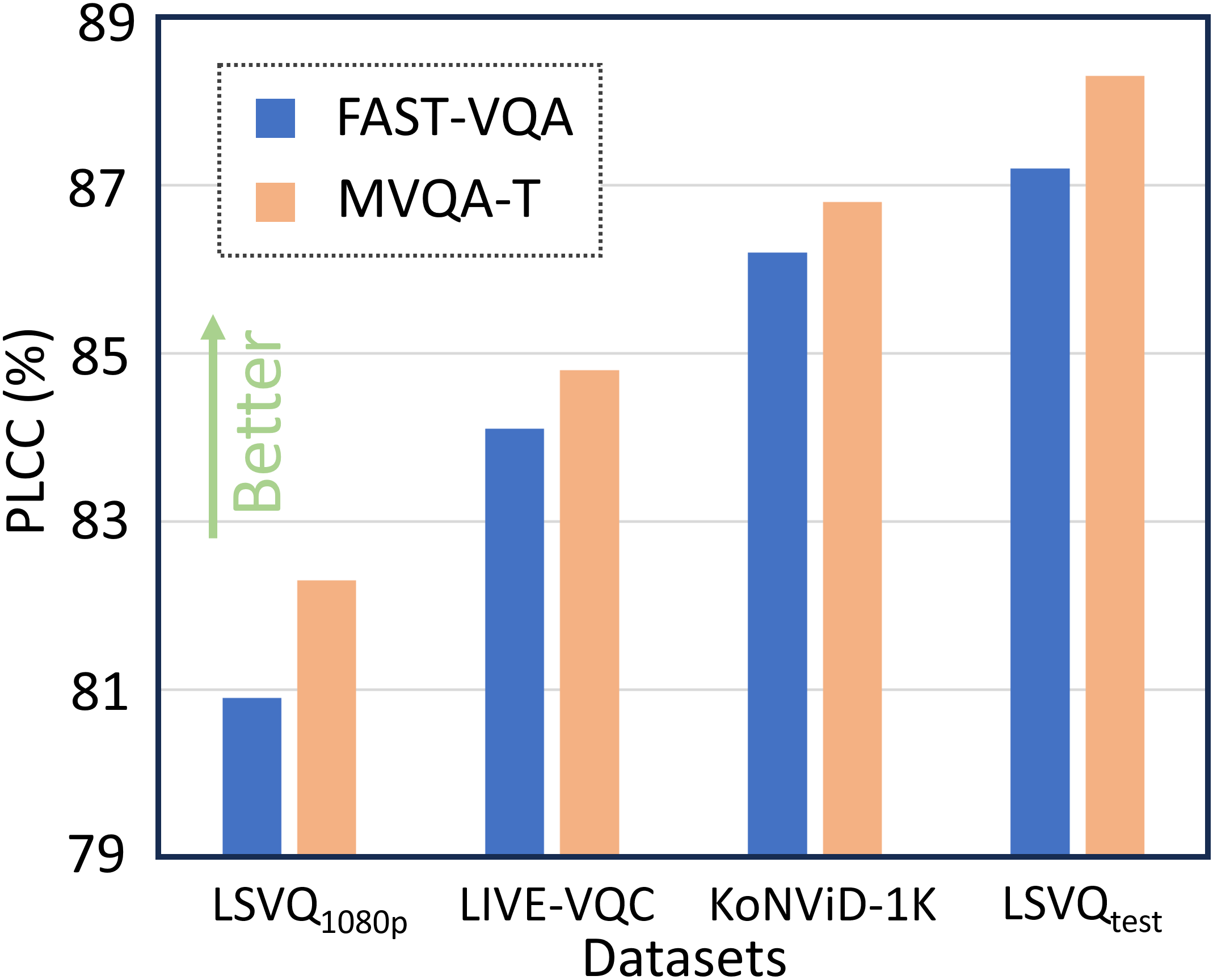}
		\centerline{\footnotesize (d) Accuracy Comparison}
	\end{minipage}
	
	\caption{Performance and efficiency comparisons between FAST-VQA\cite{paper29} and our MVQA-tiny. Our method achieves comparable results while running up to $2\times$ faster than the current most efficient FAST-VQA, and reducing GPU memory usage by $5.3\times$. This advantage further scales with increased video frames and batch size.}
	\label{fig1}
	\vspace{-0.4cm}
\end{figure}

With the widespread adoption of high-definition portable recording devices and the ongoing advancements in video compression and network technologies, ordinary users can now effortlessly capture long-duration, high-resolution videos (e.g., 2K, 4K) and upload them to online platforms.
The increasing length and size of video present substantial challenges for video quality assessment (VQA) algorithms.
Therefore, developing efficient methods to quickly evaluate massive amounts of video data to ensure a high-quality viewing experience has become a pressing challenge.

Traditional VQA methods \cite{paper7,paper8,paper9,paper10,paper11,paper12,paper13,paper14} primarily rely on handcrafted features. While these approaches are computationally efficient and have low complexity, they struggle to effectively capture the intricate, high-dimensional characteristics of videos. As a result, their performance tends to be limited when processing high-resolution videos with rich details and diverse content. 
In contrast, deep learning-based VQA methods \cite{paper41,paper15,paper42,paper16,paper17,paper18,paper29,paper23,paper43,paper19,paper45,paper46,paper20,paper68,paper69}, particularly Convolutional Neural Networks (CNNs) \cite{paper47,paper48,paper49,paper50,paper51,paper52,paper55,paper56} and Transformers \cite{paper44,paper53,paper54}, have demonstrated substantial advantages in handling complex video tasks. 
However, deep learning-based approaches typically incur high computational costs, including significant memory consumption and lengthy processing times, which become particularly challenging when dealing with large-scale, high-resolution video data. 
Specifically, although CNNs excel at extracting local features, they face limitations in modeling long-range dependencies and processing large-scale datasets. On the other hand, while Transformer models excel at capturing global dependencies, their computational complexity grows quadratically with video length and resolution, posing considerable challenges when applied to large video datasets. Consequently, relying on either CNNs or Transformers for VQA often fails to strike an effective balance between performance and computational efficiency.

Mamba \cite{paper70}, a novel sequence modeling architecture, has gained attention for addressing computational bottlenecks in processing long sequences. As a State Space Model (SSM) \cite{paper71,paper72,paper73,paper74,paper75,paper76}, Mamba models long-range dependencies with state transfer equations and linear complexity relative to input length. Beyond temporal modeling, Mamba also excels in spatio-temporal tasks \cite{paper82,paper83,paper84,paper85,paper86,paper87}, making it particularly advantageous for VQA. Traditional VQA models often struggle with redundant video data and high computational demands, but Mamba’s efficiency allows it to handle large-scale video data accurately and with reduced processing time, especially for long and high-resolution videos. Thus, Mamba-based VQA offers a practical solution for efficient and precise video quality assessment.

Moreover, the computational complexity and accuracy of VQA models are influenced not only by the model architecture but also by the sampling strategy employed. 
Traditional sampling methods like cropping \cite{paper77,paper78} and resizing \cite{paper79} reduce computational load but often distort video quality. To better retain quality information, Wu et al. \cite{paper29} introduced Grid Mini-patch Sampling (GMS), which samples small patches across uniform grids to preserve quality and temporal features. However, the fragmentation from grid sampling limits semantic information capture, which is essential for VQA accuracy \cite{paper21,paper15,paper22,paper23,paper80,paper81}. Thus, an effective sampling strategy should maintain both video quality and semantic information.

\begin{figure}[t]
	\centering
	\setlength{\abovecaptionskip}{0.1cm}
	\includegraphics[scale=0.07]{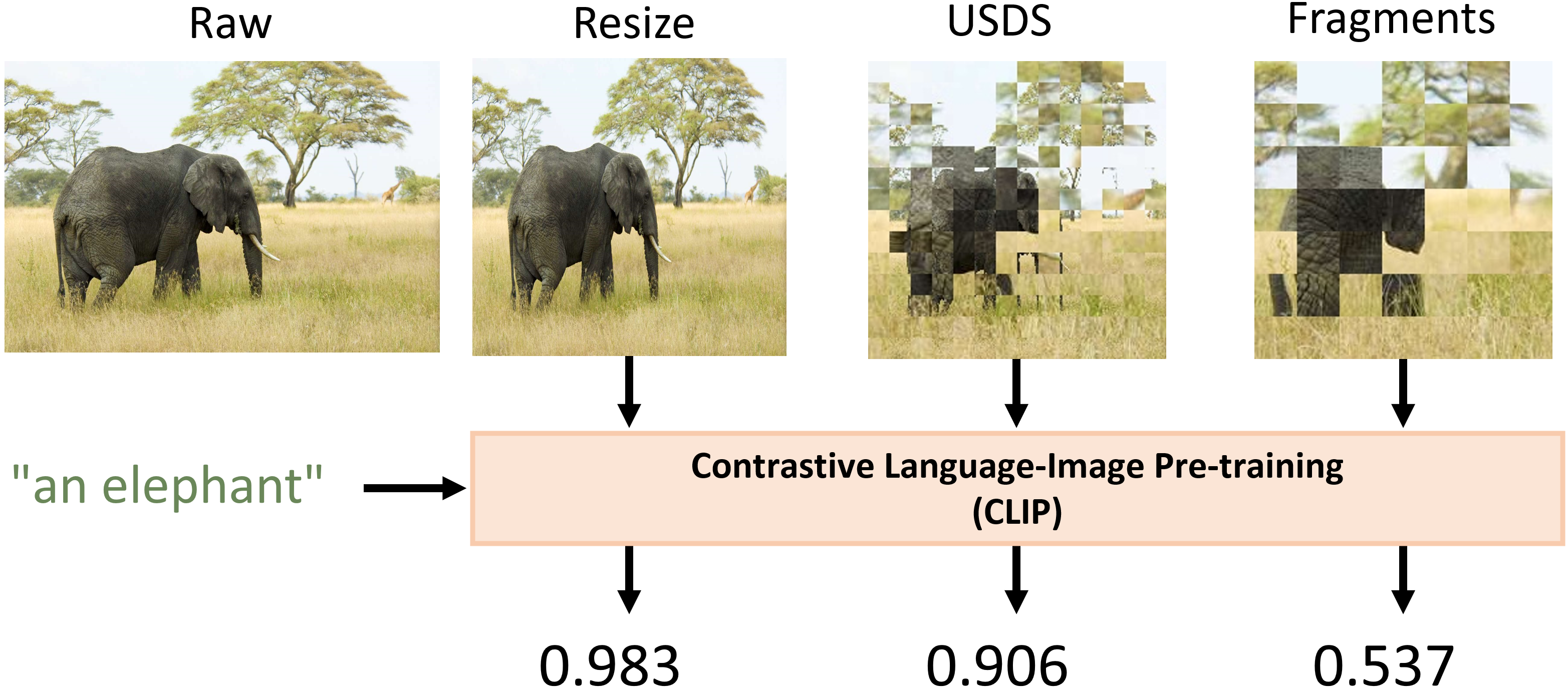}
	\caption{Comparison of semantic information retained by USDS with different samples (resize, Fragments \cite{paper29}).}
	\label{fig1.5}
	\vspace{-0.4cm}
\end{figure}

Based on the above analysis, we propose \textbf{U}nified \textbf{S}emantic and \textbf{D}istortion \textbf{S}ampling \textbf{(USDS)}, which not only effectively samples video distortions but also preserves the semantic information of the video.
As shown in \cref{fig1.5}, we use CLIP \cite{paper24}, with its strong zero-shot prediction capabilities, to verify that USDS effectively preserves semantic information.
Specifically, USDS first obtains a downsampled version to capture the semantic information of the video, then performs grid sampling to capture the distortion information. These two sampled maps are then combined at a specified ratio through masking.  Additionally, we introduce MVQA, a VQA model fully based on state space models, and experiments demonstrate that our model achieves a balance between performance and efficiency. As shown in \cref{fig1}, Compared to the current most efficient model, FastVQA, our MVQA-tiny achieves comparable performance while offering faster computation speed and lower memory usage.

Our contributions can be summarized as follows:
\begin{itemize}
	\item To the best of our knowledge, this is the first work to successfully apply state-space models in the VQA domain, which effectively balances computational efficiency and performance. 
	\item We propose a novel sampling method, USDS, which unifies semantic and distortion patch sampling through masked fusion. This approach addresses the limitation of previous sampling methods, which failed to simultaneously account for both semantic content and distortions.
	\item We conduct extensive experiments to demonstrate that MVQA outperforms both transformer-based and CNN-based methods, while ensuring efficient computation and smaller memory usage.
	
\end{itemize}

%% file: 2_related.tex
\section{Related Work}
\label{sec:formatting}

\subsection{VQA Methods}

Current VQA methods are mainly divided into two categories: traditional and deep learning-based methods.

Traditional VQA methods \cite{paper33,paper11,paper10,paper8,paper7} tend to be knowledge-driven, which evaluate the video quality by extracting hand-crafted features from spatial and temporal domains.
Earlier VQA \cite{paper11,paper33}  often extracts spatial features of video frames with the help of IQA algorithms \cite{paper30,paper31,paper32,paper38,paper39,paper40} and then predicts the video quality by feature aggregation.
In addition, VQA through natural video statistics (NVS) is also a classic class of VQA methods \cite{paper11,paper10,paper8,paper7}.
However, the performance is limited because hand-crafted features often fail to extract complex features from videos.

Deep learning-based VQA methods are primarily data-driven. Recently, numerous subjective experiments have produced large, high-quality VQA datasets \cite{paper1,paper2,paper3,paper16,paper5,paper6}, advancing deep VQA performance \cite{paper41,paper15,paper42,paper16,paper17,paper18,paper29,paper23,paper43,paper19,paper45,paper46,paper20,paper68,paper69}. Current VQA methods use either CNNs \cite{paper47,paper48,paper49,paper50,paper51,paper52,paper55,paper56} or Transformer architectures \cite{paper44,paper53,paper54}. For example, GST-VQA \cite{paper41} and VSFA \cite{paper15} employ CNNs like VGG \cite{paper47} and ResNet-50 \cite{paper48} with GRU for temporal modeling, while other studies \cite{paper16,paper18,paper22,paper43,paper62,paper42} use 3D-CNNs for spatiotemporal features. With the success of ViTs \cite{paper44}, Transformer-based VQA \cite{paper23,paper29,paper19} has gained prominence. Methods like FAST-VQA \cite{paper29} and FasterVQA \cite{paper19} sample spatial-temporal grids and utilize modified Video Swin Transformers \cite{paper53}. However, these fragment sampling strategies often neglect semantic content. Recent studies \cite{paper63,paper64,paper45,paper80} address this by adding branches for semantic extraction, such as the aesthetic branch in DOVER \cite{paper64}, a CNN branch in Zoom-VQA \cite{paper45}, and a CLIP-based branch in CLiF-VQA \cite{paper80} for aligning semantic features with linguistic prompts.

\subsection{State Space Models}
In recent years, state space models (SSMs) \cite{paper71,paper74,paper75} have emerged as one of the primary contenders to CNN and Transformer architectures within the field of deep learning.
Originally derived from classical control theory \cite{paper88}, SSMs have attracted a great deal of interest from researchers due to their great potential for remote dependency modeling and the linear growth of model complexity with respect to sequence length.
Structured state space models (S4) \cite{paper71} introduce a diagonal structure to SSMs and combine it with the diagonal plus low rank method, which improves the computational efficiency of SSMs.
Building on this, a series of studies have enhanced the S4 model to further improve its effectiveness and efficiency.
For example, S5 \cite{paper75} further introduces MIMO SSM and efficient parallel scanning in S4.
\cite{paper76} presents a new SSM layer, H3, that greatly improves the effectiveness of SSM for language modeling.
GSS \cite{paper72} builds a gated state space layer by introducing gating mechanisms in S4 to enhance its modeling capabilities.
Recently, Mamba \cite{paper70} has refined the S4 model, a data-dependent State Space Model (SSM) that incorporates efficient hardware design and a selectivity mechanism. This enhancement allows it to outperform Transformer-based models in natural language processing tasks, while also scaling linearly in complexity with input length.
Inspired by the success of mamba, a large amount of work has successfully applied mamba to a variety of vision tasks, including image understanding \cite{paper84,paper89,paper90}, video understanding \cite{paper82,paper83,paper91}, segmentation \cite{paper92,paper93}, restoration \cite{paper94,paper95,paper96}, and others \cite{paper97,paper98,paper99,paper100,paper101}. All the studies show that mamba has better performance and higher computational efficiency than CNN and Transformer models in vision tasks.

%% file: 3_approach.tex
\section{Approach}

\begin{figure*}[t]
	\centering
	\setlength{\abovecaptionskip}{0.1cm}
	\begin{subfigure}{0.30\linewidth}
		\centering
		\includegraphics[scale=0.113]{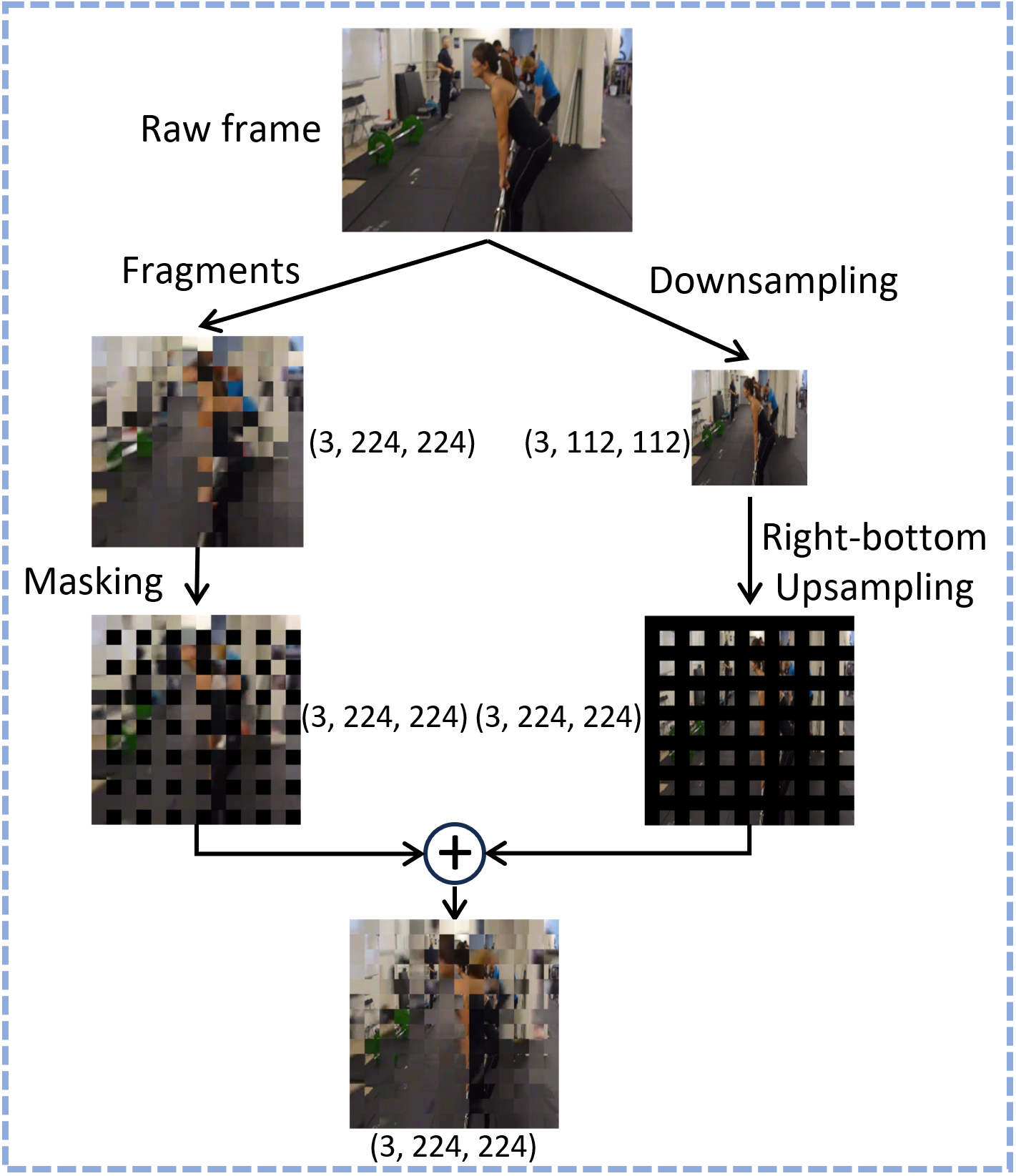}
		\caption{Unified Semantic and Distortion Sampling.}
		\label{fig3-a}
	\end{subfigure}
	\hfill
	\begin{subfigure}{0.67\linewidth}
		\centering
		\includegraphics[scale=0.113]{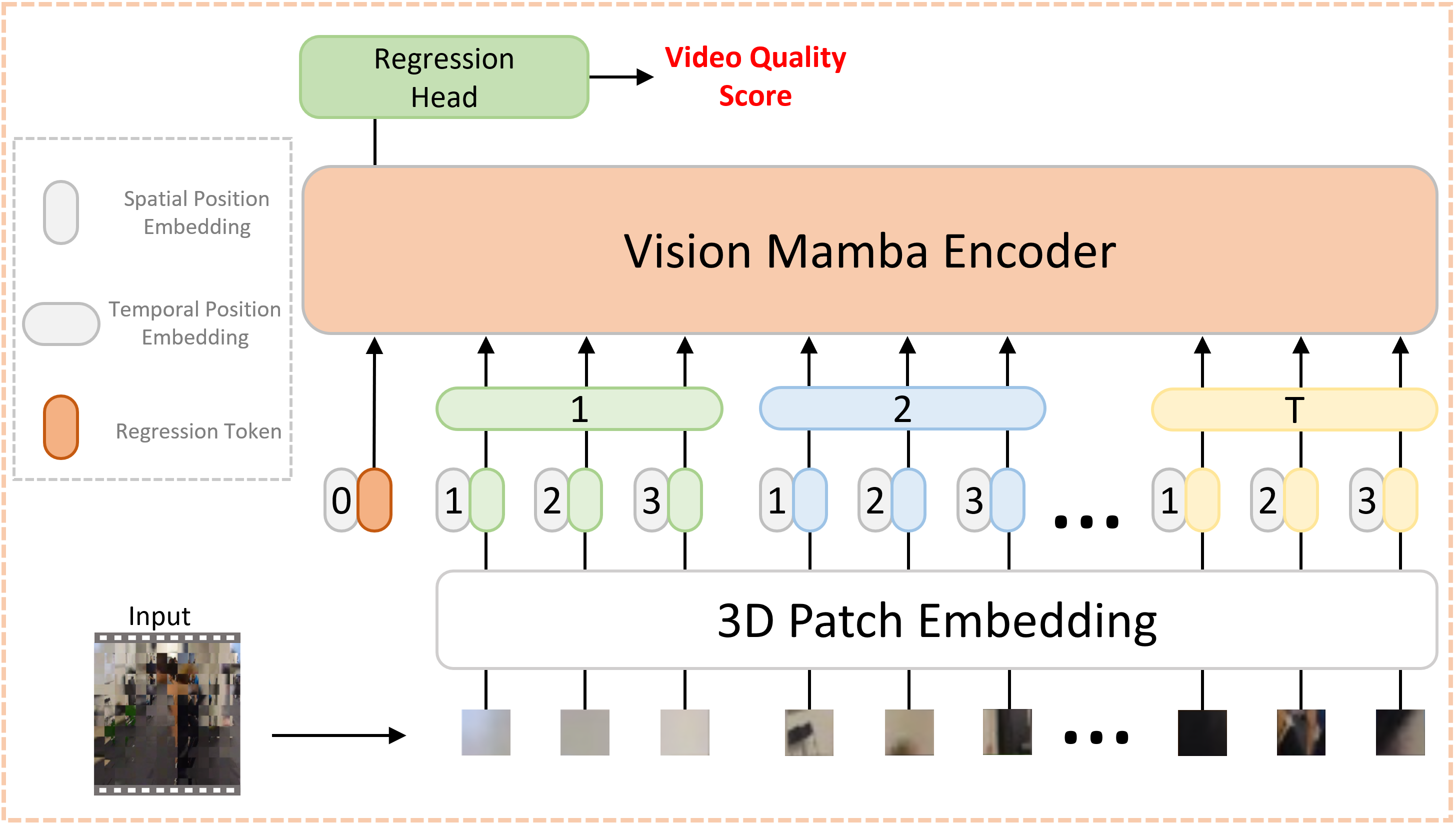}
		\caption{The network architecture of our MVQA.}
		\label{fig3-b}
	\end{subfigure}
	\caption{The framework of our proposed method. (a) The USDS consists of three distinct but interrelated stages: distortion details extraction, semantic information retention, and fusion of all resolutions. (b) MVQA transforms the input image blocks into one-dimensional vectors by 3D embedding and adds spatial location embedding and temporal location embedding to them, then extracts the features of the video using a bidirectional vision mamba encoder \cite{paper84}, and finally predicts the quality scores of the video by regressing the header. }
	\label{fig3}
	\vspace{-0.4cm}
\end{figure*}

\subsection{Preliminaries}
\textbf{State Space Models. } 
State Space Models (SSMs) are inspired by the continuous linear time-invariant (LTI) systems, which maps an input sequence $ x(t)\in \mathbb{R}^{L} $ to a latent spatial representation $ h(t)\in \mathbb{R}^{N} $ and then predicts an output sequence  $ y(t)\in \mathbb{R}^{L} $ based on that representation.
Formally, SSMs can be formulated as a linear ordinary differential equation (ODE) as follows:
\begin{align}
	h^{'}(t) &= Ah(t)+Bx(t)\\
	y(t) &= Ch(t)+Dx(t)
	\label{Eq.1}
\end{align}
where N is the dimension of the hidden state, $ h(t)\in \mathbb{R}^{N} $ is a hidden state, $ A \in \mathbb{R}^{N\times N} $ is the evolution parameter, $ B \in \mathbb{R}^{N} $ and $ C \in \mathbb{R}^{N} $ are the projection parameters,  $ D \in \mathbb{R}^{1} $ represents the skip connection.

\noindent
\textbf{Discretization.}
To process discrete inputs, the Zero-Order Holding (ZOH) rule is often utilized, which uses the sample timescale parameters $ \bigtriangleup $ to convert the continuous parameters $ A $ and $ B $ in the ODE to discrete parameters $ \overline{A} $ and $ \overline{B} $:
%
\begin{align}
	\overline{A} &= exp(\bigtriangleup A) \\
	\overline{B} &= (\bigtriangleup A) ^{-1} (exp(A)-I)\cdot \bigtriangleup B
\end{align}
where $ \overline{A}\in \mathbb{R}^{N\times N} $, $ \overline{B}\in \mathbb{R}^{N} $.
This results in the following discretized OED formula:
\begin{align}
	h_{k} &= \overline{A}h_{k-1} + \overline{B}x_{k} \\
	y_{k} &= Ch_{k} + Dx_{k}  
\end{align}

Then in order to increase the computational speed, global convolutional operations with the advantage of parallel computing are utilized to accelerate the above computational process:
\begin{align}
	\label{Eq2}
	\overline{K} &\triangleq (C{\overline{B}} ,C\overline{AB},\cdot \cdot \cdot ,C\overline{A}^{L-1}\overline{B}) \\
	y &= x \circledast \overline{K} 
\end{align}
where $ L $ is the length of the input sequence, $\circledast$ denotes convolution operation, and $\overline{K} \in \mathbb{R}^{L} $ is a structured convolutional kernel.

%

\subsection{Overall Architecture}
The architecture proposed in this paper consists of two main components: the unified semantic and distortion sampling (USDS) and the MVQA model design (\cref{fig3}). Previous studies \cite{paper29,paper19,paper102,paper23,paper103} have shown that effective sampling not only reduces model complexity but also improves performance. However, prior methods often lose critical semantic information due to fragmentation in sampled segments, affecting quality prediction. To solve this, we introduce USDS (\cref{fig3-a}), which samples video at its original resolution for distortion and downsamples for semantic information. These are fused through mask fusion to form the final sampled video. The MVQA model then uses 3D embedding on USDS results (\cref{fig3-b}), adding a regression token for quality prediction. Spatial and temporal embeddings are successively added, and the vision Mamba encoder extracts features for final quality scoring based on the regression token.

\subsection{Unified Semantic and Distortion Sampling}
\begin{figure*}[t]
	\vspace{-0.4cm}
	\centering
	\setlength{\abovecaptionskip}{0.1cm}
	\includegraphics[scale=0.15]{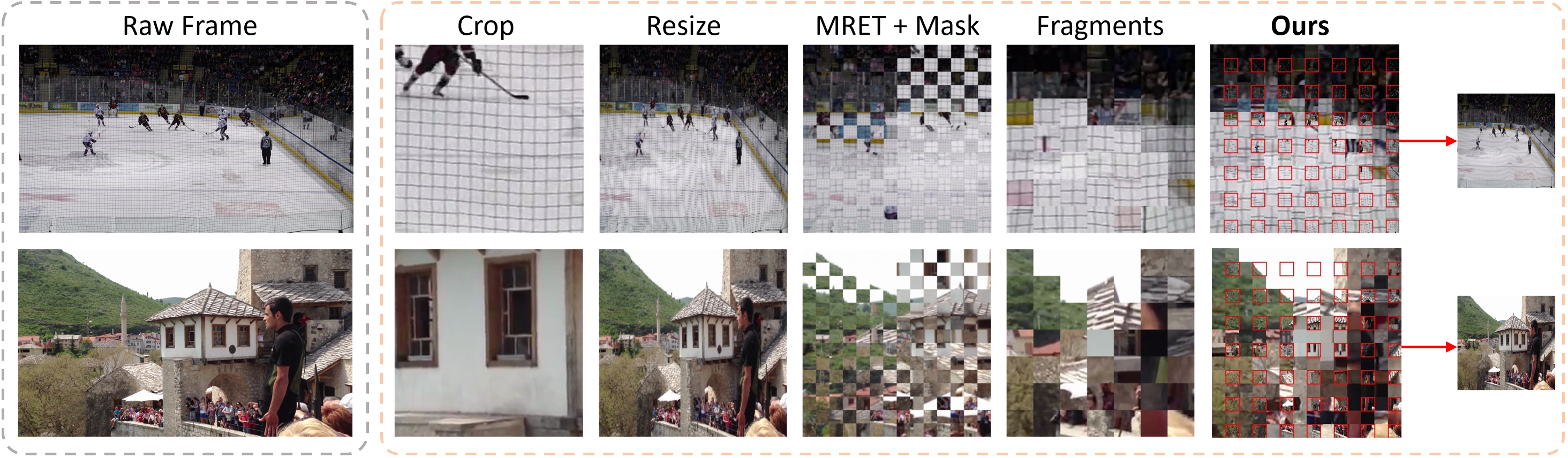}
	\caption{Comparison of USDS with crop, resize, MRET \cite{paper102}, Fragments \cite{paper29} sampling.}
	\label{fig2}
	\vspace{-0.4cm}
\end{figure*}

Considering the significant computational burden associated with model learning, we propose an approach that integrates high-resolution visual details with low-resolution semantic information to enhance the representation of video frames, which is called \textbf{U}nified \textbf{S}emantic and \textbf{D}istortion \textbf{S}ampling \textbf{(USDS)}, as shown in \cref{fig3-a}.
And the detailed USDS process is shown in \cref{alg:spatial_fragments_gms}.


\noindent
\textbf{Distortion Details and Semantic Information Extraction.}
In the first stage, input video frames $ V \in \mathbb{R}^{C \times T \times H \times W} $ are partitioned into high-resolution spatial fragments. The target resolution, denoted as $ S_h = fragments_h \times fsize_h $ and $ S_w = fragments_w \times fsize_w $ (e.g., $ 14 \times 16 = 224 $ for both), is achieved by dividing the frame into a grid of $ fragments_h \times fragments_w $ patches, each of size $ fsize_h \times fsize_w $. These fragments are sampled with random spatial offsets within each grid cell, preserving local details such as textures and edges crucial for high-fidelity reconstruction. The grid starting points are determined by evenly dividing the height and width, adjusted to avoid exceeding the frame boundaries.

For semantic information retention, each frame is downsampled to a lower resolution, typically $ S_h / 2 \times S_w / 2 $ (e.g., $ 112 \times 112 $ when $ S_h = S_w = 224 $), using bilinear interpolation:
\begin{equation}
	V_{low}[:, t, :, :] = Resize(V[:, t, :, :], (S_h / 2, S_w / 2))
\end{equation}
This process captures global contextual features, such as object boundaries and scene structure, essential for understanding the broader video context.

\noindent
\textbf{Fusion of Resolutions.}
In the second stage, the high-resolution fragments and low-resolution semantic content are fused within a multi-resolution framework. Each frame is divided into blocks of size $ 2 \cdot fsize_h \times 2 \cdot fsize_w $ (e.g., $ 32 \times 32 $ when $ fsize_h = fsize_w = 16 $), forming a grid of $ fragments_h / 2 \times fragments_w / 2 $ blocks. The high-resolution fragments are arranged to fill an intermediate frame $ \tilde{V}[:, t, :, :] $ according to their grid positions.

The low-resolution content $ V_{low} $ is expanded into a frame $ E[:, t, :, :] \in \mathbb{R}^{C \times S_h \times S_w} $ by placing each $ fsize_h \times fsize_w $ patch into the bottom-right quadrant of the corresponding $ 2 \cdot fsize_h \times 2 \cdot fsize_w $ block:
\begin{equation}
	E[:, t, R_{k,l}] = V_{low}[:, t, k' : k' + fsize_h, l' : l' + fsize_w]
\end{equation}
where $k'=0, 1, \dots, fragments_h / 2 - 1 $, and $l'=0, 1, \dots, fragments_w / 2 - 1 $, and \( R_{k,l} \) is the bottom-right $ fsize_h \times fsize_w $ region of the block with top-left corner at $ (k' \cdot 2 \cdot fsize_h, l' \cdot 2 \cdot fsize_w) $.

\begin{algorithm}[t]
	\caption{Unified Semantic and Distortion Sampling}
	\label{alg:spatial_fragments_gms}
	\begin{algorithmic}[1]
		\REQUIRE Video tensor $V \in \mathbb{R}^{C \times T \times H \times W}$, fragments $(f_h, f_w)$, size $(s_h, s_w)$, alignment $a$
		\STATE Compute $H' = f_h s_h$, $W' = f_w s_w$
		\STATE $\tilde{V}= \text{Fragments}(V, f_h, f_w, s_h, s_w, a)$
		\STATE Create mask $M \in \{0,1\}^{1 \times 1 \times H' \times W'}$:
		\[
		M_{x,y} = 
		\begin{cases} 
			1, & (x \bmod 2s_h) \ge s_h \wedge (y \bmod 2s_w) \ge s_w \\
			0, & \text{otherwise}
		\end{cases}
		\]
		\FOR{each frame $t \in [0,T)$}
		\STATE Initialize canvas: $E \leftarrow \mathbf{0}^{C \times H' \times W'}$
		\STATE $V_\text{low} \leftarrow \text{Resize}(V[:,t], \frac{H'}{2} \times \frac{W'}{2})$
		\FOR{each $(i,j)$ in $\left(\frac{H'}{2s_h}\right) \times \left(\frac{W'}{2s_w}\right)$ grid}
		\STATE Copy $E[:, 2i s_h + s_h:2(i+1) s_h, 2j s_w + s_w:2(j+1) s_w] \leftarrow V_{\text{low}}[:, i s_h:(i+1) s_h, j s_w:(j+1) s_w]$
		\ENDFOR
		\STATE $\hat{V}[:,t] \leftarrow E \odot M + \tilde{V}[:,t] \odot (1-M)$
		\ENDFOR
		\RETURN $\hat{V}$
	\end{algorithmic}
	
\end{algorithm}

A binary mask $ M \in \mathbb{R}^{1 \times 1 \times S_h \times S_w} $ is defined to select the bottom-right $ fsize_h \times fsize_w $ region of each block.
The fusion combines $ E \) and \( \tilde{V} $ for each frame $ t $:
\begin{equation}
	\hat{V}[:, t, :, :] = E[:, t, :, :] \cdot M + \tilde{V}[:, t, :, :] \cdot (1 - M)
\end{equation}
where $ \hat{V} \in \mathbb{R}^{C \times T \times S_h \times S_w} $ is the final output, blending high-resolution details in three quadrants with low-resolution semantic information in the bottom-right quadrant of each block.

The fused fragments are reassembled into complete video frames, producing the output \( \hat{V} \). \cref{fig2} illustrates a comparison between USDS and other sampling methods. By preserving fine-grained textures in high-resolution patches while integrating global semantic context from low-resolution data, USDS achieves a balanced representation that enhances both visual quality and semantic consistency.

\subsection{MVQA Model}
\cref{fig3-b} shows the structure of the proposed MVQA.
Since the mask fusion in our sampling operation is performed in terms of sample block size, we chunk the input strictly according to the sample size in order to better preserve the quality information and semantic information of the original video carried in the sampled video.
Since chunking the input destroys the spatio-temporal information of the video, in order to better model the intra-frame and inter-frame information of the input video,  we follow \cite{paper83} to add spatial position embedding for sequences within the input frames and temporal position embedding for sequences between frames.
Specifically, each input block of samples is processed using 3D convolution, where the size of the convolution kernel is $1\times 16 \times 16$. For an input video of size $X^{I}\in \mathbb{R} ^{3\times T \times H \times W} $, features of size $X^{o}\in \mathbb{R} ^{L \times C} $ are obtained after convolution, where $ L = T \times \frac{H}{16} \times \frac{W}{16}$.
Then a regression token for final quality regression and positional embedding are added to the token sequence :
\begin{equation}
	X = [ X_{reg},X ] + p_{s} + p_{t} 
\end{equation}
where $ X_{reg} $ is a learnable token that is prepended to the start of the sequence, $ p_{s}\in \mathbb{R}^{(\frac{H}{16} \times \frac{W}{16} +1)\times C}  $ is a learnable spatial position embedding, $p_{t}\in \mathbb{R}^{t\times C}  $  is a learnable temporal position embedding.

Then, we feed the sequence of tokens $ X $ into the vision mamba encoder to pass:
\begin{equation}
	T_{l} = Vim(T_{l-1} + T_{l-1}  ) 
\end{equation}
where $Vim$ represents the vision mamba block \cite{paper84} and $l$ represents the current number of layers.

Finally, we normalize the out put regression token $ X^{0}_{L} $ and feed it to the multi-layer perceptron (MLP) head to obtain the final prediction of the video quality score $ Q $ as follows:
\begin{equation}
	f = Norm(X^{0}_{L}),
	Q = MLP(f)
\end{equation}
where $ L $ denotes the number of layers of vision mamba block.

\begin{figure}[t]
	\centering
	\setlength{\abovecaptionskip}{0.1cm}
	\includegraphics[scale=0.17]{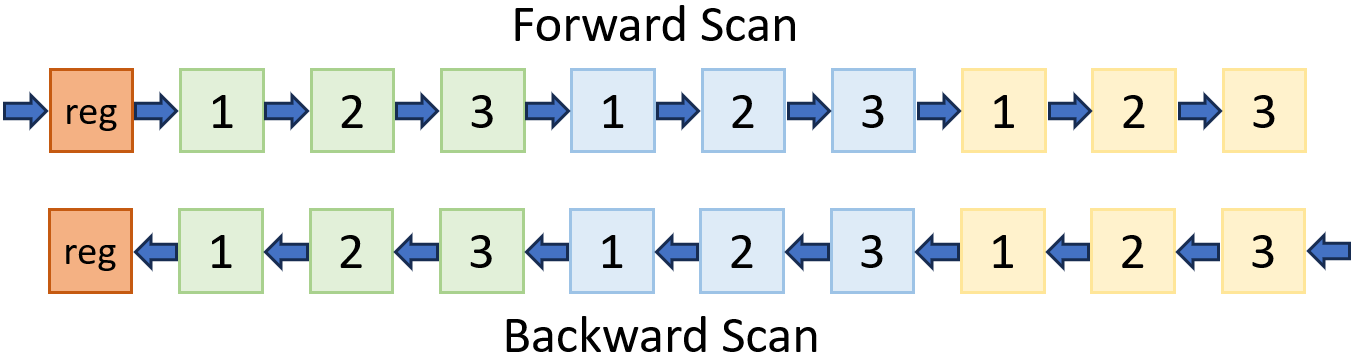}
	\caption{Scan method.}
	\label{fig4}
	\vspace{-0.4cm}
\end{figure}

Vision Mamba \cite{paper84} introduces a bidirectional scanning mechanism that processes tokens forward and backward, enhancing Mamba's spatial modeling and yielding strong results on image tasks. However, video data requires both spatial and temporal modeling. Li et al. \cite{paper83} found that a spatial-first, temporal-next scanning approach is most effective for video tasks, so MVQA adopts this mechanism, as shown in \cref{fig4}.

\begin{table*}[h]
	\vspace{-0.4cm}
	\setlength{\abovecaptionskip}{0.01cm} 
	\caption{Experimental performance of the pre-trained MVQA model on the LSVQ dataset on four test sets (LSVQ$_{test}$, LSVQ$_{1080p}$, KoNViD-1k, LIVE-VQC). 
		Existing best multi-branch in \textbf{\textcolor{green!50!black}{green}} and existing best single-branch in \textbf{\textcolor{blue}{blue}}.}
	\centering
	\label{table1}
	\small
	
	\setlength{\aboverulesep}{0pt}
	\setlength{\belowrulesep}{0pt}
	\renewcommand\arraystretch{0.9}
	\centering\setlength{\tabcolsep}{5.1pt}
	
	\begin{tabular}{c|c|c|cc|cc|cc|cc}
		\toprule
		\multicolumn{3}{c|}{Testing Type} & \multicolumn{4}{c|}{Intra-dataset Test Datasets} & \multicolumn{4}{c}{Cross-dataset Test Datasets} \\
		\hline
		\multicolumn{3}{c|}{Testing Datasets} & \multicolumn{2}{c|}{\textbf{LSVQ$_{test}$}} & \multicolumn{2}{c|}{\textbf{LSVQ$_{1080p}$}} & \multicolumn{2}{c|}{\textbf{KoNViD-1k}} & \multicolumn{2}{c}{\textbf{LIVE-VQC}} \\
		\hline
		Type & Methods & Source & SROCC & PLCC & SROCC & PLCC & SROCC & PLCC & SROCC & PLCC \\
		\hline
		\multirow{3}{*}{Classical} & BRISQUE \cite{paper30} &\footnotesize \itshape TIP, 2012 &  0.569 & 0.576  & 0.497  & 0.531  & 0.646  & 0.647  & 0.524  & 0.536 \\ 
		& TLVQM \cite{paper8} &\footnotesize \itshape TIP, 2019 & 0.772 & 0.774  & 0.589  & 0.616  & 0.732  & 0.724  & 0.670  & 0.691 \\ 
		& VIDEVAL \cite{paper7} &\footnotesize \itshape TIP, 2021 &  0.794 & 0.783  & 0.545  & 0.554  & 0.751  & 0.741  & 0.630  & 0.640 \\ 
		\hdashline
		\multirow{6}{*}{\shortstack{Deep \\ Multi-branch}} 
		& PVQ$_{wo/patch}$ \cite{paper16} & \footnotesize\itshape CVPR, 2021 &  0.814 & 0.816  & 0.686  & 0.708  & 0.781  & 0.781  & 0.747  & 0.776 \\ 
		& PVQ$_{w/patch}$ \cite{paper16} &\footnotesize\itshape  CVPR, 2021 &  0.827 & 0.828  & 0.711  & 0.739  & 0.791  & 0.795  & 0.770  & 0.807 \\
		& BVQA \cite{paper17} & \footnotesize\itshape TCSVT, 2022 &  0.852 & 0.854  & 0.771  & 0.782  & 0.834  & 0.837  & 0.816  & 0.824 \\
		& DOVER  \cite{paper64}& \footnotesize\itshape ICCV, 2023 &  0.881 & 0.879  & 0.782  & 0.827  & 0.871  & 0.872  & 0.812  & 0.841 \\
		& Zoom-VQA \cite{paper45} & \footnotesize\itshape CVPR,2023 & 0.886 & 0.879 & 0.799  & 0.819 & 0.877 & 0.875 & 0.814 & 0.833 \\
		& Q-Align \cite{paper105} & \footnotesize\itshape ICML, 2024 & 0.883  & 0.882  & 0.797  & 0.830  & 0.865  & 0.877  & NA  & NA \\
		& CLiF-VQA \cite{paper80}  & \footnotesize\itshape ACMMM,2024 &  0.886 & 0.887  & 0.790  & 0.832  & 0.877  & 0.874  & \textbf{\textcolor{green!50!black}{0.834}}  & \textbf{\textcolor{green!50!black}{0.855}} \\
		& MBVQA \cite{paper106} & \footnotesize\itshape CVPR, 2024 & \textbf{\textcolor{green!50!black}{0.895}}  & \textbf{\textcolor{green!50!black}{0.895}}  & \textbf{\textcolor{green!50!black}{0.809}}  & \textbf{\textcolor{green!50!black}{0.844}}  & \textbf{\textcolor{green!50!black}{0.878}}  & \textbf{\textcolor{green!50!black}{0.884}}  & 0.806  & 0.844 \\
		
		\hdashline
		\multirow{6}{*}{\shortstack{Deep \\ Single-branch}} & VSFA \cite{paper15}& \footnotesize\itshape ACMMM, 2019 & 0.801 & 0.796 & 0.675  & 0.704  & 0.784  & 0.794  & 0.734  & 0.772 \\ 
		& FAST-VQA-M \cite{paper29}& \footnotesize\itshape ECCV, 2022 &  0.852 & 0.854  & 0.739  & 0.773  & 0.841  & 0.832  & 0.788  & 0.810 \\
		& FAST-VQA \cite{paper29}& \footnotesize\itshape ECCV, 2022 &  0.872 & \textbf{\textcolor{blue}{0.874}}  & 0.770  & 0.809  & \textbf{\textcolor{blue}{0.864}}  & 0.862  & \textbf{\textcolor{blue}{0.824}}  & \textbf{\textcolor{blue}{0.841}} \\
		
		& FasterVQA \cite{paper19}  & \footnotesize\itshape TPAMI, 2023 &  \textbf{\textcolor{blue}{0.873}} & \textbf{\textcolor{blue}{0.874}}  & \textbf{\textcolor{blue}{0.772}}  & \textbf{\textcolor{blue}{0.811}}  & 0.863  & \textbf{\textcolor{blue}{0.863}}  & 0.813  & 0.837 \\
		
		
		\cline{2-11}
		& \textbf{MVQA-tiny }  & \textbf{\itshape ours} &  \textbf{0.882} & \textbf{0.883}  & \textbf{0.781}  & \textbf{0.823}  & \textbf{0.870}  & \textbf{0.868}  & \textbf{0.828}  & \textbf{0.848} \\
		& \textbf{MVQA-middle}   & \textbf{\itshape ours} &  \textbf{\textcolor{red}{0.898}} & \textbf{\textcolor{red}{0.899}}  & \textbf{\textcolor{red}{0.812}}  & \textbf{\textcolor{red}{0.846}}  & \textbf{\textcolor{red}{0.885}}  & \textbf{\textcolor{red}{0.887}}  & \textbf{\textcolor{red}{0.852}}  & \textbf{\textcolor{red}{0.873}} \\
		\hdashline
		\multicolumn{3}{c|}{\footnotesize \itshape improvement to existing best Single-branch}  &\footnotesize \itshape \textcolor{blue}{$ 2.86\%$} &\footnotesize \itshape \textcolor{blue}{$ 2.86\%$}  &\footnotesize \itshape \textcolor{blue}{$ 5.18\%$}  &\footnotesize \itshape \textcolor{blue}{$ 4.32\%$} &\footnotesize \itshape \textcolor{blue}{$ 2.43\%$} &\footnotesize \itshape \textcolor{blue}{$ 2.78\%$} &\footnotesize \itshape \textcolor{blue}{$ 3.40\%$} &\footnotesize \itshape \textcolor{blue}{$ 3.80\%$} \\
		\hdashline
		\multicolumn{3}{c|}{\footnotesize \itshape improvement to existing best Multi-branch}  &\footnotesize \itshape \textcolor{green!50!black}{$ 0.34\% $} &\footnotesize \itshape \textcolor{green!50!black}{$ 0.45\% $}  &\footnotesize \itshape \textcolor{green!50!black}{$ 0.37\% $}  &\footnotesize \itshape \textcolor{green!50!black}{$ 0.24\% $} &\footnotesize \itshape \textcolor{green!50!black}{$ 0.80\% $} &\footnotesize \itshape \textcolor{green!50!black}{$ 0.34\% $} &\footnotesize \itshape \textcolor{green!50!black}{$ 2.16\% $} &\footnotesize \itshape \textcolor{green!50!black}{$ 2.11\% $} \\
		\bottomrule
	\end{tabular}
	\vspace{-0.4cm}
\end{table*}

\subsection{Loss Function}
Several studies have shown that both monotonicity-induced loss and linearity-induced loss play a very important role in the quality prediction task, so we combine both as our loss function.
Given the predicted quality score $ \hat{Q}=\left \{ \hat{q_{1}},\hat{q_{2}},...,\hat{q_{m}} \right \} $ and ground-truth subjective quality score $ Q=\left \{ q_{1},q_{2},...,q_{m} \right \} $.

The monotonicity-induced loss $ L_{mon} $ and the linearity-induced loss $ L_{lin} $ are defined as follows:
\begin{equation}
	L_{mon} =\frac{1}{m^{2} } \sum_{i=1}^{m} \sum_{j=1}^{m}max(0,\left | q_{i}- q_{j} \right |-f(q_{i}, q_{j})\cdot (\hat{q_{i}}-\hat{q_{j}} ) ) 
\end{equation}
where $ f(q_{i}, q_{j})=1 $ if $ q_{i}\ge q_{j} $, otherwise $ f(q_{i}, q_{j})=-1 $.

\begin{equation}
	L_{lin} = ( 1-\frac{ {\textstyle \sum_{i=1}^{m}(\hat{q}_{i}-\hat{a} )({q}_{i}-{a})} }{\sqrt{ {\textstyle \sum_{i=1}^m{(\hat{q}_{i}- \hat{a})^{2} {\textstyle \sum_{i=1}^{m}(q_{i}-a)^{2}}  }} } } )/2
\end{equation}
where $ a= \frac{1}{m} {\textstyle \sum_{i=1}^{m}} q_{i} $ and $ \hat{a} = \frac{1}{m} {\textstyle \sum_{i=1}^{m}} \hat{q}_{i} $.

The total loss function $ L $ is obtained by combining the two loss functions $ L_{mon} $ and $ L_{lin} $ above:
\begin{equation}
	L=\alpha L_{mon}+\beta L_{lin}
\end{equation}
where $ \alpha $ and $ \beta $ represent the weights of monotonicity-induced loss and linearity-induced loss.

%% file: 4_experiments.tex
\section{Experiments}
\label{sec:5}

\subsection{Experimental Setups}

\noindent
\textbf{Datasets.} 
We test the model on four datasets including LSVQ \cite{paper16}, KoNViD-1k (1200 videos) \cite{paper1}, LIVE-VQC (585 videos) \cite{paper3}, and YouTube-UGC (1067 videos) \cite{paper6}. 
Specifically, we pre-train MVQA on LSVQ$_{train}$, a subset of  LSVQ containing 28,056 videos. 
Intra-dataset testing is performed on two subsets of LSVQ, LSVQ$_{test}$ (7400 videos) and LSVQ$_{1080p}$ (3600 videos).
We perform cross-dataset testing on KoNViD-1k and LIVE-VQC.
Further, we fine-tune the model on  KoNViD-1k, LIVE-VQC, and YouTube-UGC.
It should be noted that  YouTube-UGC contains 1500 videos, but only 1067 videos are available to us.

\noindent
\textbf{Evaluation Criteria.} Spearman Rank Order Correlation Coefficient (SROCC) and Pearson Linear Correlation Coefficient (PLCC) are used as evaluation Metrics. 
Specifically, SRCC is used to measure the prediction monotonicity between predicted scores and true scores by ranking the values in both series and calculating the linear correlation between the two ranked series. 
In contrast, PLCC evaluates prediction accuracy by calculating the linear correlation between a series of predicted scores and true scores.
And higher SROCC and PLCC scores indicate better performance.

\noindent
\textbf{Implementation Details.} 
We employ PyTorch framework and an NVIDIA H100 card to train the model.
We pre-train the MVQA backbone network on the Kinetics-400  \cite{paper65} dataset.
We set the initial learning rate to 0.0025, the optimizer to AdamW, and use a cosine annealing strategy to dynamically adjust the learning rate. 
The detailed information of our proposed MVQA with two different scales is shown in \cref{table0}.
For MVQA-tiny and MVQA-middle, we set the batch size to 32 and 8, and the weight decay during training to 0.1 and 0.05.

\begin{table}[t]
	\setlength{\abovecaptionskip}{0.01cm}
	\caption{Details of MVQA with different sizes.}
	\centering\setlength{\tabcolsep}{4.7pt}
	\label{table0}
	\footnotesize
	\setlength{\aboverulesep}{0pt}
	\setlength{\belowrulesep}{0pt}

	\begin{tabular}{l|c|c|c|c|c}
		\toprule
		Models & Input Size& Depth & Dim & Param & Flops   \\
		\hline
		MVQA-tiny & $32\times224 \times 224$ & 24 & 192 & 7M & 34G    \\
		MVQA-middle & $32\times224 \times224$ & 32 & 576 & 74M & 403G   \\
		\bottomrule
	\end{tabular}
	\vspace{-0.4cm}
\end{table}

\begin{table*}[t]
	\vspace{-0.4cm}
	\setlength{\abovecaptionskip}{0.01cm} 
	\caption{The finetune results on LIVE-VQC, KoNViD and YouTube-UGC.
		Existing best multi-branch in \textbf{\textcolor{green!50!black}{green}} and existing best single-branch in \textbf{\textcolor{blue}{blue}}.
	}
	\label{table2}
	\small
	\setlength{\aboverulesep}{0pt}
	\setlength{\belowrulesep}{0pt}
	\renewcommand\arraystretch{0.9}
	\centering\setlength{\tabcolsep}{4pt}
	
	\begin{tabular}{c|c|c|cc|cc|cc|cc}
		\toprule
		\multicolumn{3}{c|}{Finetune Datasets} & \multicolumn{2}{c|}{\textbf{LIVE-VQC}} & \multicolumn{2}{c|}{\textbf{KoNViD-1k}} & \multicolumn{2}{c|}{\textbf{YouTube-UGC}} & \multicolumn{2}{c}{\textbf{\itshape Average}}  \\
		\hline
		Type & Methods & Source & SROCC & PLCC & SROCC & PLCC & SROCC & PLCC & SROCC & PLCC   \\
		\hline
		\multirow{3}{*}{Classical} & TLVQM \cite{paper8}& \footnotesize \itshape TIP, 2019 & 0.799 & 0.803  & 0.773  & 0.768  & 0.669  & 0.659 & 0.732 & 0.726 \\ 
		& VIDEVAL \cite{paper7} & \footnotesize \itshape TIP, 2021 &  0.752 & 0.751  & 0.783  & 0.780  & 0.779  & 0.773 & 0.772 & 0.772 \\ 
		& RAPIQUE \cite{paper66} & \footnotesize \itshape OJSP, 2021 &  0.755 & 0.786  & 0.803  & 0.817  & 0.759  & 0.768 & 0.774 & 0.790 \\
		\hdashline 
		\multirow{2}{*}{Classical +  Deep} & CNN+TLVQM \cite{paper67} & \footnotesize \itshape ACMMM, 2020 & 0.825 & 0.834 & 0.816 & 0.818 & 0.809 & 0.802 & 0.815 & 0.814\\
		& CNN+VIDEVAL \cite{paper7} & \footnotesize \itshape TIP, 2021 & 0.785 & 0.810 & 0.815 & 0.817 & 0.808 & 0.803 & 0.806 & 0.810\\
		\hdashline
		\multirow{6}{*}{\shortstack{Deep \\ Multi-branch}} 
		& PVQ \cite{paper16} &  \footnotesize \itshape CVPR, 2021 &  0.827 & 0.837  & 0.791  & 0.786  & NA  & NA  & NA & NA \\ 
		& BVQA \cite{paper17} & \footnotesize \itshape TCSVT, 2022 &  0.831 & 0.842  & 0.834  & 0.836  & 0.831  & 0.819 & 0.832 & 0.832 \\
		& CoINVQ \cite{paper22} & \footnotesize \itshape TCSVT, 2021 & NA & NA & 0.767 & 0.764 & 0.816 & 0.802 & NA & NA \\
		& DOVER \cite{paper64} & \footnotesize \itshape ICCV, 2023 &  0.812 & 0.852  & 0.897  &0.899  & 0.877  & 0.873 & 0.862 & 0.875\\
		& MaxVQA \cite{paper63} &\footnotesize \itshape ACMMM,2023 & 0.854 & 0.873 & 0.894 & 0.895 & \textbf{\textcolor{green!50!black}{0.894}} & \textbf{\textcolor{green!50!black}{0.890}} & 0.881 & 0.886 \\
		& CLiF-VQA \cite{paper80} & \footnotesize \itshape ACMMM,2024 &  \textbf{\textcolor{green!50!black}{0.866}} & 0.878  & \textbf{\textcolor{green!50!black}{0.903}}  & 0.903  & 0.888  & \textbf{\textcolor{green!50!black}{0.890}}  & \textbf{\textcolor{green!50!black}{0.886}} & \textbf{\textcolor{green!50!black}{0.890}} \\
		& MBVQA \cite{paper106} & \footnotesize \itshape CVPR, 2024 &  0.860 & \textbf{\textcolor{green!50!black}{0.880}}  & 0.901  & \textbf{\textcolor{green!50!black}{0.905}}  & 0.876  & 0.877 & 0.879 & 0.887\\
		\hdashline
		\multirow{7}{*}{\shortstack{Deep \\ Single-branch}} & VSFA \cite{paper15}& \footnotesize \itshape ACMMM, 2019 & 0.773 & 0.795 & 0.773  & 0.775  & 0.724  & 0.743 & 0.752 & 0.765 \\ 
		& GST-VQA \cite{paper41} & \footnotesize \itshape TCSVT, 2021 & NA & NA & 0.814 & 0.825 & NA & NA & NA & NA \\ 
		
		& FAST-VQA-M \cite{paper29}& \footnotesize \itshape ECCV, 2022 &  0.803 & 0.828  & 0.873  & 0.872  & 0.768  & 0.765 & 0.815 & 0.822 \\
		& FAST-VQA \cite{paper29} & \footnotesize \itshape ECCV, 2022 &  \textbf{\textcolor{blue}{0.845}} & 0.852  & 0.890  & 0.889  &  0.857  & 0.853  & 0.864 & 0.865\\
		& FasterVQA \cite{paper19}  & \footnotesize \itshape TPAMI, 2023 &  0.843 & \textbf{\textcolor{blue}{0.858}}  & \textbf{\textcolor{blue}{0.895}}  & \textbf{\textcolor{blue}{0.898}}  & \textbf{\textcolor{blue}{0.863}}  & \textbf{\textcolor{blue}{0.859}}  & \textbf{\textcolor{blue}{0.867}} & \textbf{\textcolor{blue}{0.872}}\\
		\cline{2-11}
		& \textbf{MVQA-tiny} &  \textbf{\itshape ours} & \textbf{0.850} & \textbf{0.866} & \textbf{0.903} & \textbf{0.904} & \textbf{0.869} & \textbf{0.872} & \textbf{0.874} & \textbf{0.881} \\ 
		& \textbf{MVQA-middle} & \textbf{ \itshape ours} & \textbf{\textcolor{red}{0.878}} & \textbf{\textcolor{red}{0.895}} & \textbf{\textcolor{red}{0.925}} & \textbf{\textcolor{red}{0.925}} & \textbf{\textcolor{red}{0.901}} & \textbf{\textcolor{red}{0.903}} & \textbf{\textcolor{red}{0.901}} & \textbf{\textcolor{red}{0.908}} \\ 
		\hdashline
		\multicolumn{3}{c|}{\footnotesize \itshape improvement to existing best Single-branch}  &\footnotesize \itshape \textcolor{blue}{$3.91 \%$} &\footnotesize \itshape \textcolor{blue}{$ 4.31\%$}  &\footnotesize \itshape \textcolor{blue}{$ 3.35\%$}  &\footnotesize \itshape \textcolor{blue}{$ 3.01\%$} &\footnotesize \itshape \textcolor{blue}{$ 4.40\%$} &\footnotesize \itshape \textcolor{blue}{$ 5.12\%$} &\footnotesize \itshape \textcolor{blue}{$ 3.92\%$} &\footnotesize \itshape \textcolor{blue}{$ 4.13\%$} \\
		\hdashline
		\multicolumn{3}{c|}{\footnotesize \itshape improvement to existing best Multi-branch}  &\footnotesize \itshape \textcolor{green!50!black}{$ 1.39\% $} &\footnotesize \itshape \textcolor{green!50!black}{$ 1.71\% $}  &\footnotesize \itshape \textcolor{green!50!black}{$ 2.44\% $}  &\footnotesize \itshape \textcolor{green!50!black}{$ 2.21\% $} &\footnotesize \itshape \textcolor{green!50!black}{$ 0.78\% $} &\footnotesize \itshape \textcolor{green!50!black}{$ 1.46\% $} &\footnotesize \itshape \textcolor{green!50!black}{$ 1.69\% $} &\footnotesize \itshape \textcolor{green!50!black}{$ 2.02\% $} \\
		\bottomrule
	\end{tabular}
	\vspace{-0.4cm}
\end{table*}

\subsection{Experiment Results}

\textbf{Pre-training Results on LSVQ.}
We pre-train the proposed MVQA on LSVQ and conduct intra-dataset testing on LSVQ$_{test}$ and LSVQ$_{1080p}$. Additionally, cross-dataset testing was performed on KoNViD-1k and LIVE-VQC. 
The results are shown in \cref{table1}.
First, for our tiny version of the MVQA model, compared to the current best-performing and fastest FasterVQA model, our model achieves comparable performance while offering faster processing speed.
In addition, for the middle version of MVQA, compared with the classical methods  (BRISQUE, TLVQM, VIDEVAL), our method shows superior performance on all test sets. Our model also achieves better results compared to deep learning-based methods. 
Specifically, compared to the current state-of-the-art single-branch models, our model achieve average improvements of $4.02 \%$ and $ 3.59 \%$ in SROCC and PLCC metrics for intra-dataset testing. In cross-dataset testing, the improvements are $2.92 \%$ and $ 3.29 \%$, respectively.
Since single-branch models often struggle to simultaneously capture both quality and semantic features of videos, recent multi-branch models, which better incorporate video semantic information, have shown outstanding performance.
The performance of our single-branch based MVQA model is still ahead of all current multi-branch models.
Compared to the current state-of-the-art results, our model achieves improvements of $0.92 \%$  and $ 0.79 \%$ in SROCC and PLCC, respectively, while offering higher computational efficiency.

\noindent
\textbf{Fine-tuning Results on Small Datasets.}
\begin{table}[h]
	\setlength{\abovecaptionskip}{0.01cm}
	\caption{FLOPs and running time(average of 10 runs) on GPU (RTX 3090). FLOPs and time are in $G$ and $s$, respectively.}
	\centering
	\footnotesize
	\setlength\tabcolsep{3.9pt}  
	\renewcommand\arraystretch{1}
	
	\setlength{\aboverulesep}{0pt}
	\setlength{\belowrulesep}{0pt}
	\begin{tabular}{l|c:c|c:c|c:c}
		\toprule
		\multirow{2}{*}{Methods} & \multicolumn{2}{c|}{\textbf{540p}} & \multicolumn{2}{c|}{\textbf{720p}} & \multicolumn{2}{c}{\textbf{1080p}} \\
		\cline{ 2 - 7}
		&FLOPs & Time &  FLOPs & Time &FLOPs & Time  \\
		\hline
		VSFA \cite{paper15} & 6440 & 1.506  & 11426 & 2.556 & 25712 & 5.291\\
		PVQ \cite{paper16} & 9203 &1.792&13842&2.968&36760&6.556\\
		BVQA \cite{paper17} &17705 &3.145&31533&7.813&70714 &14.34\\
		FAST-VQA \cite{paper29} & 284 &0.056&284&0.056 &284&0.056\\
		FasterVQA \cite{paper19} & 70  & 0.032  & 70 & 0.032 & 70  & 0.032 \\
		DOVER \cite{paper64} &282 &0.061&282&0.061&282&0.062 \\
		CLiF-VQA \cite{paper80} & 1432 &0.522&1432&0.522&1432&0.522\\
		MBVQA \cite{paper106} & 912 & 0.194 & 1232 & 0.267 & 2150 & 0.891 \\
		\hline
		\textbf{MVQA-tiny} & \textbf{34} & \textbf{0.028}& \textbf{34} & \textbf{0.028} &\textbf{34} &\textbf{0.028}\\
		\textbf{MVQA-middle} &\textbf{ 403} & \textbf{0.130} &\textbf{ 403} & \textbf{0.130} & \textbf{403} & \textbf{0.130} \\
		\bottomrule
	\end{tabular}
	\label{table3}
	\vspace{-0.4cm}
\end{table}
We fine-tune MVQA on three small datasets (LIVE-VQC, KoNViD-1k, YouTube-UGC), as shown in \cref{table2}.
We can see that MVQA-tiny achieves comparable performance compared to the current best single-branch model.
Moreover, MVQA-middle achieves improvements of $3.92 \%$ and $4.13 \%$ in SROCC and PLCC, respectively, on three datasets compared to the best results from single-branch models.
Additionally, compared to the best multi-branch models, it shows improvements of  $1.69 \%$ and  $2.02 \%$ in SROCC and PLCC, respectively.


\subsection{Efficiency}
To test the efficiency, we compare MVQA with the current mainstream deep learning-based models.
Specifically, we compare the FLOPs and GPU runtimes for videos of different resolutions, where the length of the videos are 150 frames, as shown in \cref{table3}.
Compared to CNN-based models (VSFA, PVQ, BVQA), MVQA-tiny reduces FLOPs by up to $756\times$, $1081\times$, and $2079\times$, as well as reduces computation time by up to $189\times$, $234\times$, and $512\times$, respectively.
Moreover, compared to the fastest FasterVQA, our model has comparable computational computation as well as a $2\times $ reduction in flops.
In addition, MVQA-middle not only performs better but also reduce FLOPs by up to $3.55\times $ and $5.33 \times$, and computation time by up to $ 4.02\times$ and $6.85 \times$, respectively, compared to the current best models CLiF-VQA and MBVQA.

\begin{table}[t]
	\vspace{-0.4cm}
	\setlength{\abovecaptionskip}{0.01cm}
	\caption{USDS performance in the  Video Swin Transformer.}
	\centering\setlength{\tabcolsep}{3.1pt}
	\label{table12}
	\footnotesize
	\setlength{\aboverulesep}{0pt}
	\setlength{\belowrulesep}{0pt}
	\begin{tabular}{l|cc|cc|cc}
		\toprule
		Datasets &\multicolumn{2}{c|}{\textbf{LSVQ$_{test}$}} & \multicolumn{2}{c|}{\textbf{KoNViD-1k}} & \multicolumn{2}{c}{\textbf{LIVE-VQC}}   \\
		\hline
		Sampling & SROCC & PLCC & SROCC & PLCC & SROCC & PLCC  \\
		\hline
		\itshape Fragments & 0.872 & 0.874 & 0.864 & 0.862 & 0.824 & 0.841 \\
		\textbf{USDS} & \textcolor{red}{0.877} & \textcolor{red}{0.879} & \textcolor{red}{0.871} & \textcolor{red}{0.869} & \textcolor{red}{0.832} & \textcolor{red}{0.851} \\
		\bottomrule
	\end{tabular}
\end{table}

\subsection{Evaluation on USDS}

\begin{figure}[t]
	\vspace{-0.2cm}
	\centering
	\setlength{\abovecaptionskip}{0.1cm}
	\includegraphics[scale=0.3]{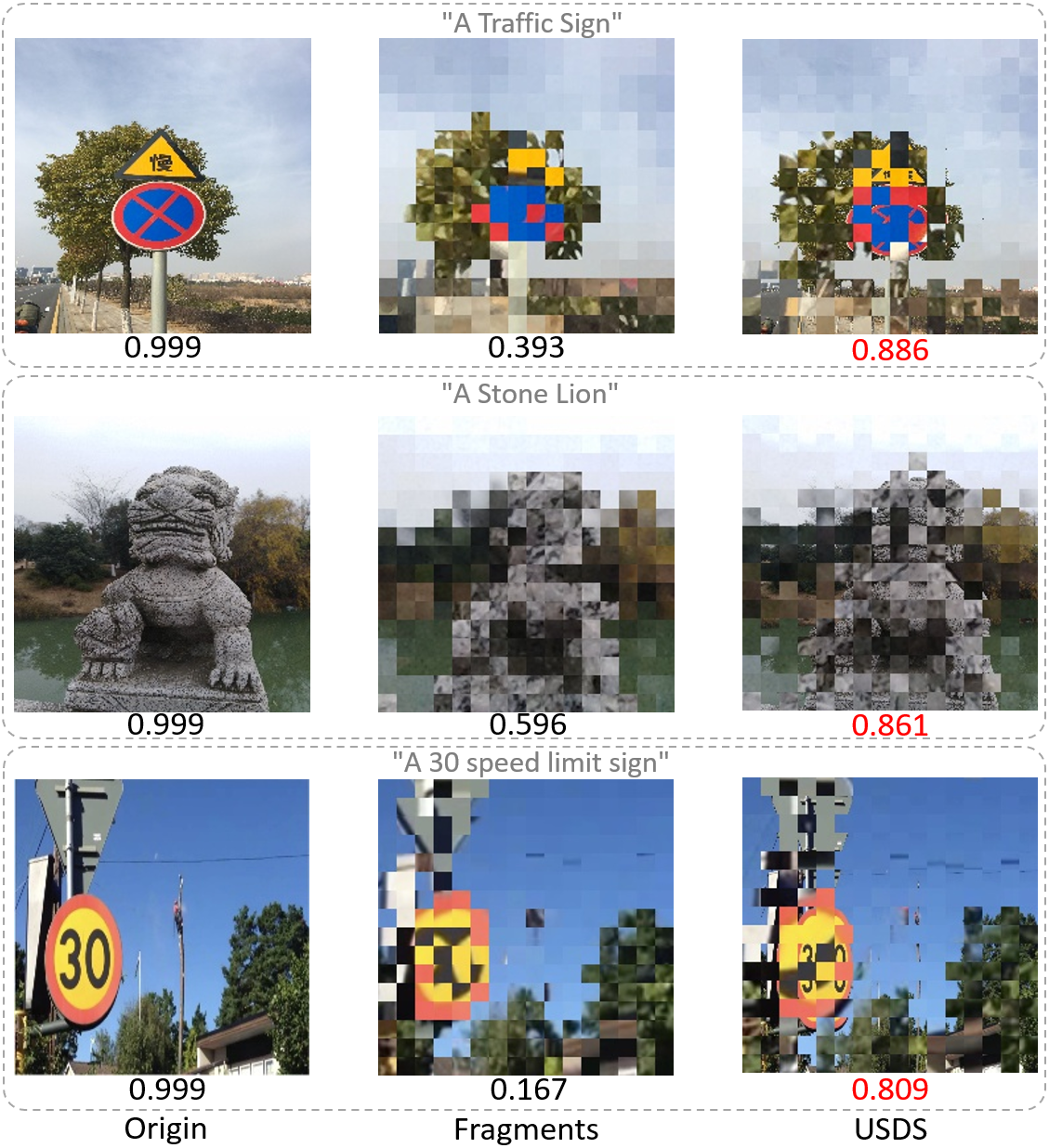}
	\caption{Results of semantic analysis experiments.}
	\label{qualitative analysis}
	\vspace{-0.4cm}
\end{figure}

\noindent
\textbf{Semantic Analysis.}
We conduct experiments to verify that USDS can capture sufficient semantic information.
Consistent with the testing method in Fig. \ref{fig1.5}, 
we use CLIP model with strong visual language capabilities to validate the semantic information in the sampled maps.
As shown in Fig. \ref{qualitative analysis}, USDS can retain more semantics compared to Fragments.
More qualitative results can be found in the Appendix.



\noindent
\textbf{Cross-Architecture Validation.}
To demonstrate the advantages of USDS under different models.
We validate USDS's effectiveness in the Video Swin Transformer, the dominant architecture in VQA, as shown in Tab. \ref{table12}.
The results show that USDS is not only effective in the proposed MVQA architecture, but also in the Video Swin Transformer model due to the traditional Fragments sampling method.

\subsection{Ablation Studies}


\noindent
\textbf{Ablation on Sampling.} 
We compare the proposed USDS with four common sampling methods (resize, crop, MRET \cite{paper102}, fragments \cite{paper29}) to validate its effectiveness. 
To ensure a fair comparison, we keep the model structure and the block size used for frame segmentation during input unchanged. As shown in \cref{table4}.
Specifically, USDS performs very well compared to traditional resize and crop sampling.
Additionally, compared to the currently most effective sampling method, fragments , which can fully and evenly preserve the local texture of the video, USDS shows improvements of $2.99 \%$ and $2.00 \%$ in SROCC and PLCC, respectively. This demonstrates that USDS not only preserves the local texture of the video to the greatest extent but also retains more semantic information.

\begin{table}[t]
	\vspace{-0.4cm}
	\setlength{\abovecaptionskip}{0.01cm}
	\caption{Ablation study on different sampling methods in MVQA.}
	\centering\setlength{\tabcolsep}{3.3pt}
	\label{table4}
	\footnotesize
	\setlength{\aboverulesep}{0pt}
	\setlength{\belowrulesep}{0pt}
	\begin{tabular}{l|cc|cc|cc}
		\toprule
		Datasets &\multicolumn{2}{c|}{\textbf{LSVQ$_{test}$}} & \multicolumn{2}{c|}{\textbf{KoNViD-1k}} & \multicolumn{2}{c}{\textbf{LIVE-VQC}}   \\
		\hline
		Sampling & SROCC & PLCC & SROCC & PLCC & SROCC & PLCC  \\
		\hline
		\itshape Resize & 0.847 & 0.851 & 0.834 & 0.836 & 0.762 & 0.809  \\
		\itshape Crop & 0.798 & 0.810 & 0.765 & 0.779 & 0.744 & 0.762  \\
		\itshape MRET \cite{paper102} & 0.851 &0.854 & 0.841 & 0.839 & 0.785 & 0.820  \\
		\itshape Fragments \cite{paper29} & 0.864 & 0.865 & 0.850 & 0.853& 0.791 & 0.829 \\
		\hdashline
		\textbf{USDS} & \textcolor{red}{0.882} & \textcolor{red}{0.883} & \textcolor{red}{0.870} & \textcolor{red}{0.868} & \textcolor{red}{0.828} & \textcolor{red}{0.848} \\
		\bottomrule
	\end{tabular}
	\vspace{-0.2cm}
\end{table}

\begin{figure}[t]
	\centering
	\setlength{\abovecaptionskip}{0.1cm}
	\begin{minipage}{0.49\linewidth}
		\centering
		\includegraphics[width=1\linewidth]{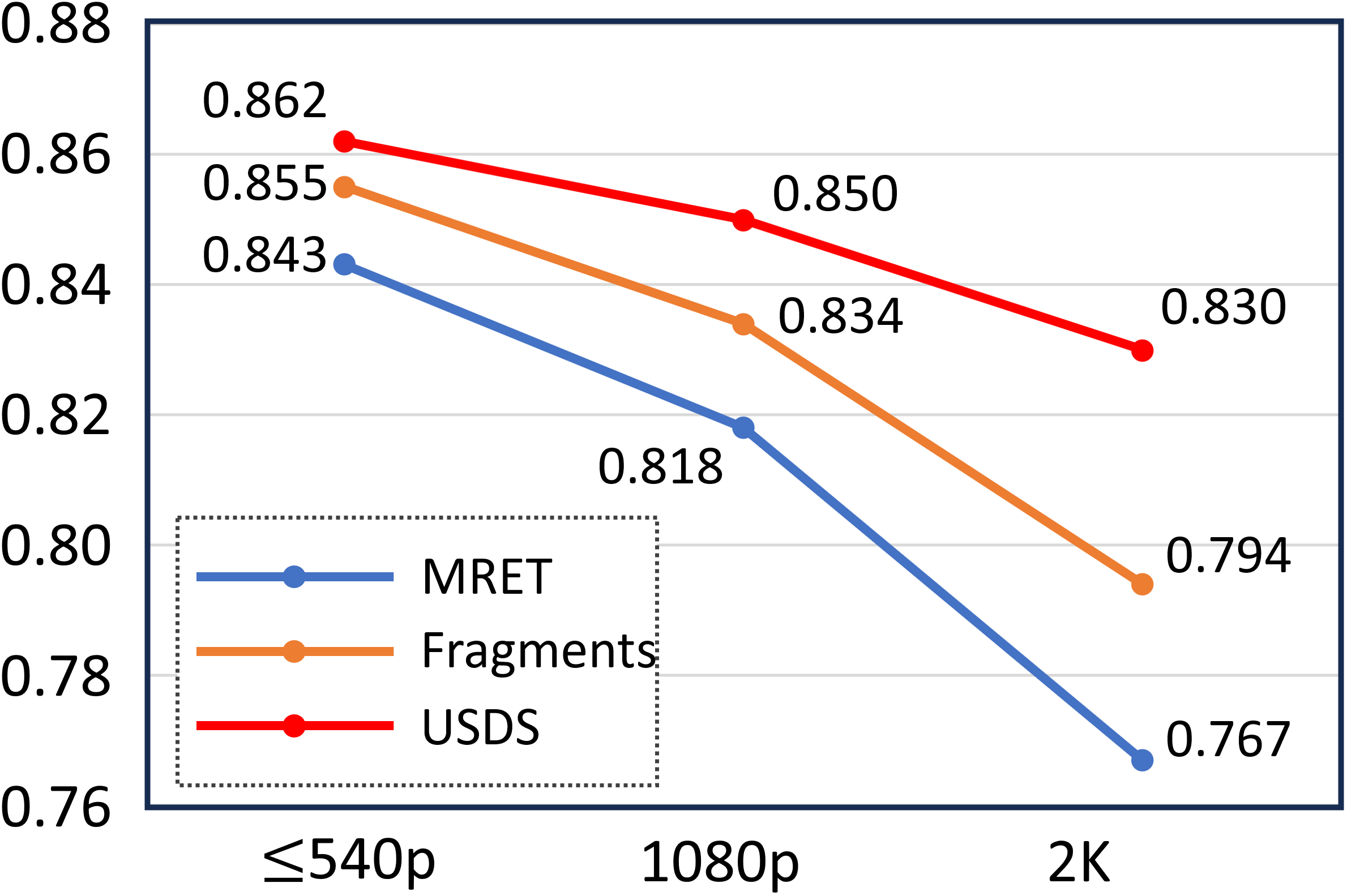}
		\centerline{ \footnotesize (a) PLCC}
	\end{minipage}
	\begin{minipage}{0.49\linewidth}
		\centering
		\includegraphics[width=1\linewidth]{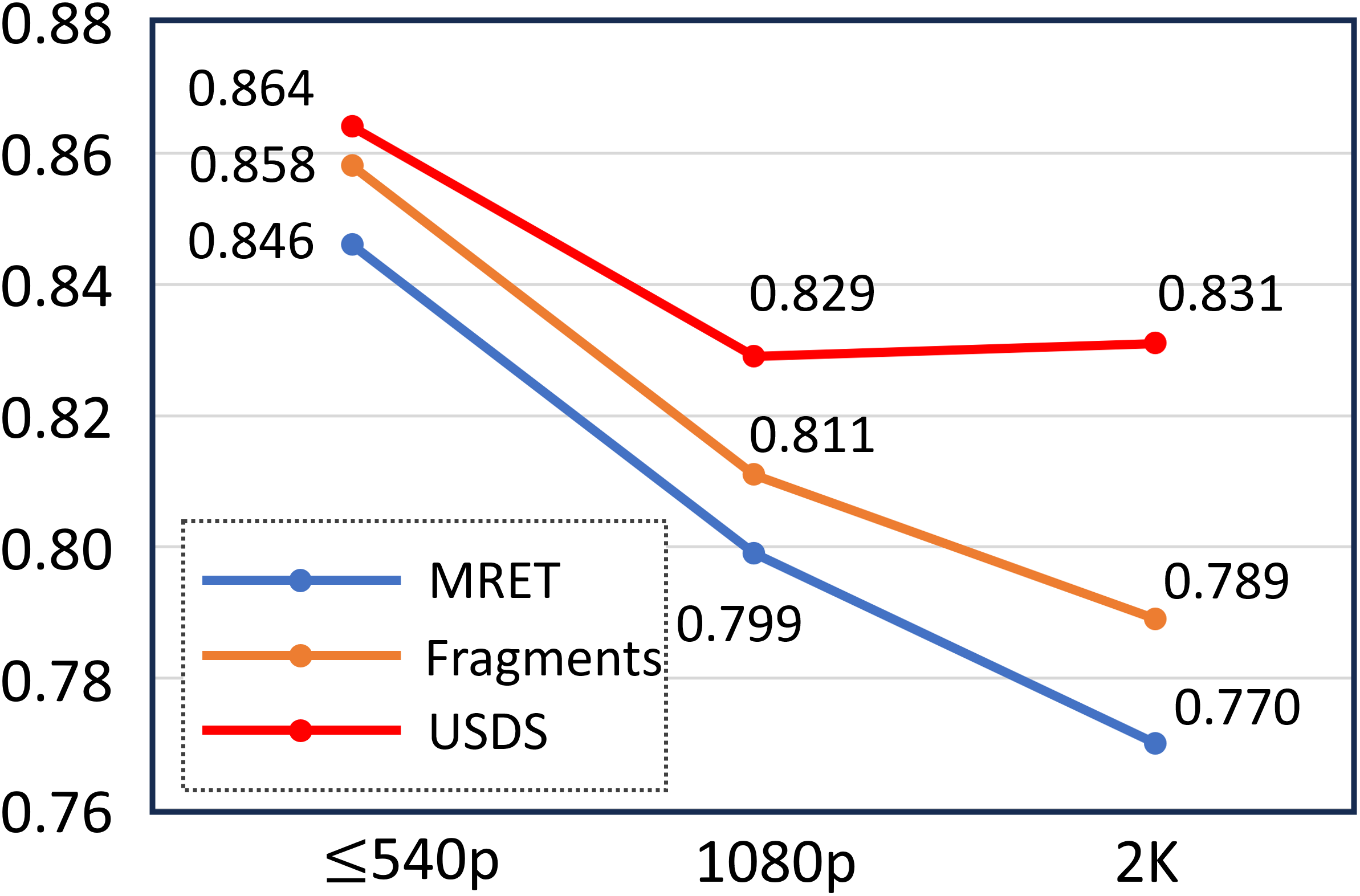}
		\centerline{\footnotesize (b) SROCC}
	\end{minipage}
	\caption{Ablation study on different resolutions. }
	\label{fig6}
	\vspace{-0.4cm}
\end{figure}

\noindent
\textbf{Ablation on Different Resolutions.} 
To further validate the effectiveness of the semantic information extracted by USDS, we conducted ablation experiments at different video resolutions. Since both MRET \cite{paper102} and Fragments \cite{paper29} sample fixed-size blocks from the video, at higher video resolutions, the sampling results become highly fragmented, often leading to a complete loss of semantic information.
As shown in \cref{fig6}.
Since the MRET sampling strategy performs multi-scale sampling from four regions of the video, when the video resolution is too high, it can only sample from localized areas of the video. This results in the sampling not fully capturing the quality information of the video.
Although the Fragments sampling method can sample uniformly from the video, at high resolutions, the features sampled using this method completely lose semantic information.
The experimental results show that as the resolution increases, the advantages of USDS become more pronounced. This is because we have effectively embedded the semantic information during the sampling process.

%% file: 5_conclusion.tex
\section{Conclusion}
In this paper, we are the first to explore the application of recent state-space models (Mamba) in VQA, aiming to balance computational efficiency with excellent performance. We first propose an effective sampling method, USDS, which efficiently integrates video semantic information, and introduce MVQA, a VQA model entirely based on state-space models (SSM). Extensive experiments across multiple datasets demonstrate that MVQA is a simple, effective, and highly efficient VQA model.